\documentclass[sigconf,screen]{acmart}

\copyrightyear{2026}
\acmYear{2026}
\setcopyright{cc}
\setcctype{by}
\acmConference[MM '26] {Proceedings of the 35th ACM International Conference on Multimedia}{November 10--14, 2026}{Rio de Janeiro, Brazil.}
\acmBooktitle{Proceedings of the 35th ACM International Conference on Multimedia (MM '26), November 10--14, 2026, Rio de Janeiro, Brazil}
\acmISBN{979-8-4007-2213-4/2026/11}
\acmDOI{10.1145/3767308.3836298}

\renewcommand{\shortauthors}{He et al.}

\setcitestyle{numbers,sort&compress}
\usepackage{svg}
\usepackage[dvipsnames]{xcolor}
\usepackage{enumitem}
\usepackage{latexsym}
\usepackage[T1]{fontenc}
\usepackage[utf8]{inputenc}
\usepackage{bbding}
\usepackage{microtype}
\usepackage{amsmath}

\usepackage{booktabs}
\usepackage{multirow}
\usepackage{colortbl}
\usepackage{makecell}
\usepackage{graphicx}
\usepackage{arydshln}
\usepackage{float}      
\usepackage{caption}    
\usepackage{subcaption}
\usepackage{verbatim} 
\usepackage{tcolorbox} 
\tcbuselibrary{breakable} 
\usepackage{array}
\usepackage{geometry}   
\usepackage{pgfplots}
\pgfplotsset{compat=1.18}
\usepgfplotslibrary{groupplots}
\usepackage{hyperref}
\usepackage[normalem]{ulem}

\newcommand{\flaw}[1]{\textcolor{purple}{#1}}
\newcommand{\good}[1]{\textcolor{Green}{#1}}

\newtcolorbox{calloutblock}{
    colback=white, 
    colframe=black, 
    boxrule=0.5pt, 
    arc=0pt, 
    left=1pt, 
    right=1pt, 
    top=1pt, 
    bottom=1pt, 
    breakable,
    fontupper=\footnotesize,
}

\begin{document}

\title{Novel Claim or D\'ej\`a Vu? Rethinking ``Contamination-Free'' Dynamic Evaluation for Multimodal Automated Fact-Checking}

\author{Haorui He}
\affiliation{%
  \institution{Department of Interactive Media, Hong Kong Baptist University}
  \country{}
  }
\affiliation{%
  \institution{School of Computing and Data Science, The University of Hong Kong}
  \country{}
  }
\email{harryhe@connect.hku.hk}

\author{Xinwen Chen}
\affiliation{%
  \institution{Faculty of Science and Technology, Beijing Normal-Hong Kong Baptist University}
  \country{}
  }
\email{xinwwwc@gmail.com}

\author{Dacheng Wen}
\affiliation{%
  \institution{Department of Interactive Media, Hong Kong Baptist University}
  \country{}
  }
\affiliation{%
  \institution{School of Computing and Data Science, The University of Hong Kong}
  \country{}
  }
\email{wdacheng@connect.hku.hk}

\author{Reynold Cheng}
\affiliation{%
  \institution{School of Computing and Data Science, The University of Hong Kong}
  \country{}
}
\email{ckcheng@cs.hku.hk}

\author{Francis C. M. Lau}
\affiliation{%
  \institution{School of Computing and Data Science, The University of Hong Kong}
  \country{}}
\email{fcmlau@cs.hku.hk}

\author{Yupeng Li}
\authornote{Corresponding author. This work was done while Haorui He was under the supervision of Yupeng Li.}
\affiliation{%
  \institution{Department of Interactive Media, Hong Kong Baptist University}
  \country{}
  }
\email{ivanypli@gmail.com}

\begin{abstract}
Multimodal automated fact-checking (MAFC) verifies claims by retrieving and reasoning over external evidence. 
However, most existing static benchmarks risk contamination: 
they primarily consist of outdated claims verifiable using an LLM's internal knowledge without external evidence.
This can inflate performance estimates and fail to reflect true capability on novel claims that require up-to-date information.
To address this, emerging dynamic benchmarks collect claims published after LLMs' knowledge cut-off dates, assuming they are uncontaminated.
This work revisits this assumption by empirically studying contamination risks in both the state-of-the-art (SOTA) static AVeriTeC benchmark and our newly constructed dynamic ClaimReview2025Q4 benchmark, as well as their impact on MAFC evaluation. Our experiments yield 16 findings, highlighting three key results: 
(1) Dynamic evaluation reduces but does not eliminate contamination risks, as 17.09\%--29.30\% of post-cut-off claims remain potentially contaminated; 
(2) Many newly published claims can be verified either directly or by synthesizing multiple pieces of public knowledge available before the cut-off; and 
(3) Contamination can induce statistically significant inflation in MAFC performance, increasing Macro-F1 by up to 11.34 points and distorting system rankings.
In light of these findings, we re-evaluate SOTA LLMs under a strictly contamination-controlled setting. 
Our study provides practical guidelines for trustworthy MAFC evaluation.
\end{abstract}



\maketitle

\section{Introduction}
\label{sec:intro}
The rapid and widespread dissemination of (multimodal) misinformation poses a pressing societal challenge that cannot be addressed by professional human fact-checkers alone~\cite{factcheck_definition, www_mcfend, guo2022survey, multimodal_survey, mdfend}, 
which motivates automated countermeasures. 
State-of-the-art (SOTA) multimodal automated fact-checking (MAFC) systems, e.g., DEFAME \cite{defame}, leverage powerful multimodal large language models (LLMs) to retrieve and analyze textual and/or visual evidence from online sources (e.g., the Web) to verify (check-worthy) claims.\footnote{Check-worthy claims hold significant public interest and affect public behavior \cite{annotation_schema}.} 

\begin{figure}[t]
\centering 
\includegraphics[width=\linewidth]{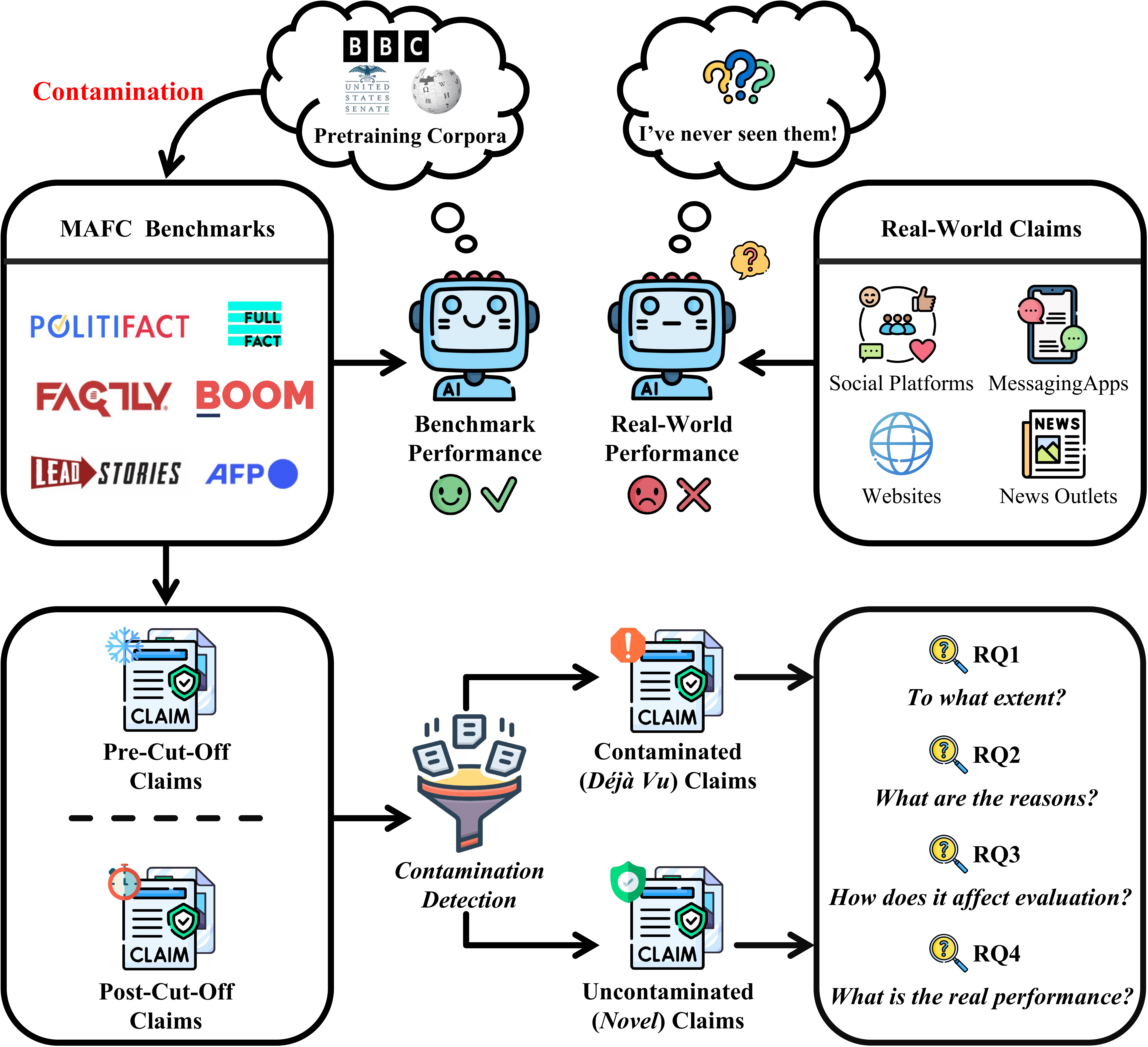}
\caption{An overview of contamination in MAFC and our studied research questions (RQs).}
\label{fig:overview}
\end{figure}

The evaluation of MAFC systems relies on benchmarks built from claims drawn from fact-checking articles published by professional agencies.
However, these benchmarks, such as AVeriTeC~\cite{averitec}, are \emph{static} and are not updated after their initial construction.
Over time, their outdated coverage of claims from past events introduces a risk of \emph{contamination}: 
the LLM backbones powering the MAFC systems have seen these events and related context during pretraining, 
enabling them to verify claims from parametric knowledge alone and to sidestep the core MAFC challenge of retrieving and reasoning over external evidence~\cite{xfacta, factcheck_definition}.\footnote{Table~\ref{tab:rq2_cases} provides examples of such contaminated claims.}
As illustrated in Fig.~\ref{fig:overview}, 
\textbf{such contamination can artificially inflate estimated MAFC performance relative to true capability on genuinely novel claims arising from timely news events that fall outside the LLMs' training data}~\cite{xfacta, factcheck_definition}.
To address this, recent benchmarks such as VERITAS~\cite{veritas} and XFACTA~\cite{xfacta} adopt a ``\emph{dynamic}'' evaluation paradigm.
They continuously aggregate claims published after models' knowledge cut-off dates and regularly refresh the dataset (e.g., monthly or quarterly), aiming to keep the evaluation data unseen and support reliable assessment of MAFC systems.

\textbf{However, four critical issues remain unresolved.}
\emph{First, the extent of contamination in existing static MAFC benchmarks is unknown.}
Prior studies have not directly quantified the contamination risks in these benchmarks, making it difficult to assess the reliability of their evaluation results.
\emph{Second, it is unclear how effectively dynamic benchmarks eliminate contamination risks.}
Claims published after LLMs' knowledge cut-off dates may still be verifiable using information available before the cut-off, or may revisit previously fact-checked topics~\cite{fact-checked}. 
Thus, some seemingly ``novel'' claims may in fact be \emph{déjà vu} to LLMs. 
As a result, 
dynamic benchmarks could still overestimate MAFC performance on truly unseen claims, undermining the goal of uncontaminated evaluation.
These contamination risks in both static and dynamic benchmarks raise a \emph{third} issue: \emph{How does such contamination distort MAFC evaluation?}
While \citet{veritas} observed performance declines on claims published after LLMs' knowledge cut-off dates, empirical evidence based on grouping claims by these dates remains mixed: \citet{quelle2024perils} report no abrupt drop, and \citet{fontana2025evaluating} find declines only for real claims. 
Moreover, temporal performance shifts may stem from factors unrelated to contamination, such as distributional shifts~\cite{www_mcfend} or changes in claim complexity~\cite{HLE, browsecomp}.
Consequently, without directly isolating and controlling for contamination, it is unclear to what extent existing benchmark scores may be artificially inflated.
\emph{Finally}, given the contamination risks in existing benchmarks and their potential to distort evaluation, \emph{the true MAFC capabilities of SOTA LLMs remain unclear.} Without benchmarks that rigorously exclude claims verifiable via pre-cut-off knowledge, we cannot assess models' generalization to genuinely unseen claims from fast-evolving, real-world media environments.

To systematically investigate these issues, this work conducts empirical studies to address the following research questions (RQs).
\begin{itemize}[leftmargin=*]
\item 
\textbf{RQ1: To what extent are existing static and dynamic MAFC benchmarks contaminated?} 
To address RQ1, we adopt an established evidence sufficiency evaluation pipeline proposed in AVeriTeC~\cite{averitec} for detecting potential contamination.
Specifically, we measure the sufficiency of evidence extracted directly from an LLM's internal knowledge against the oracle evidence used by human fact-checkers for the same claim, flagging high-sufficiency claims as potentially contaminated.
We evaluate six SOTA LLMs, including both proprietary models (e.g., \textit{GPT-5.2} and \textit{Gemini-3.0-Pro}) and open-source models (e.g., \textit{DeepSeek-V3.2} and \textit{Qwen3.5-122B-A10B}).
We then compare the SOTA static MAFC benchmark, AVeriTeC\footnote{
According to the manual evaluation by \citet{multimodal_survey}, 28.68\% of claims in AVeriTeC either contain multimodal content or require multimodal reasoning to be fact-checked.}, with ClaimReview2025Q4, our newly constructed benchmark comprising claims published in Q4 2025 to simulate post-cut-off evaluation.
As shown in Sec.~\ref{sec:rq1}, dynamic evaluation reduces but does not eliminate contamination: 17.09\%--29.30\% of claims remain potentially contaminated.

\item 
\textbf{RQ2: How does contamination arise in dynamic benchmarks?}
To address RQ2, 
we conduct a case study of claims whose oracle fact-checking articles were published after LLMs' knowledge cut-off dates, yet whose LLM-generated articles still show high similarity to those oracle articles.
As shown in Sec.~\ref{sec:rq2}, 
such contamination persists because many newly published claims can be verified using public knowledge available before the cut-off, either directly or by synthesizing multiple known facts.
\item 
\textbf{RQ3: How does contamination affect MAFC performance evaluation?}
For RQ3, we compare Accuracy and Macro-F1 of the six LLMs on contaminated vs. uncontaminated claim subsets.
As shown in Sec.~\ref{sec:rq3}, contamination can significantly inflate MAFC performance by as much as 11.34 percentage points in Macro-F1 and can even alter model rankings. 
Further analysis reveals that contamination enables the LLMs to achieve higher retrieval precision, bypassing exploration of the evidence space.
\item 
\textbf{RQ4: How do SOTA LLMs perform under contamination-controlled MAFC evaluation?} 
Building on the findings from RQ1--RQ3, we re-evaluate all models on a unified contamination-controlled set containing only claims that are uncontaminated for all six models under all three similarity metrics. As detailed in Sec.~\ref{sec:clean_eval}, DeepSeek-V3.2 achieves the highest Macro-F1 score. 
However, all models score below 56\% Macro-F1, underscoring substantial room for improvement in MAFC solutions.
\end{itemize}

Our code and appendix are available at \url{https://trustworthycomp.github.io/Rethink-MAFC-Eval/}.

\section{Related Work}\label{sec:related}
 
Automated fact-checking (AFC) systems verify check-worthy claims by retrieving relevant evidence and predicting a verdict, such as \textit{Supported}, \textit{Refuted}, or \textit{Not Enough Evidence}, often accompanied by an explanatory justification.\footnote{Definitions of these verdict categories are provided in Appendix~\ref{app:verdict}.}
The majority of AFC systems are text-based~\cite{guo2022survey, averitec, averitec_challenge, he2025debating, fact2fiction, hero, luo2024message, bertemo}. 
However, multimodal misinformation, which combines text with other modalities such as images or videos, has been shown to be even more misleading~\cite{mm_mislead1, mm_mislead2, Ragar, fakesv, fakett, fakevv}. Thus, recent research has increasingly focused on MAFC systems, typically built on multimodal LLMs. For example, RAGAR~\cite{Ragar} leverages LLMs to generate textual descriptions for images to enable multimodal understanding, 
while DEFAME~\cite{defame} employs LLMs to dynamically select tools for extracting and reasoning over both textual and visual evidence.

Most existing systems are evaluated on static benchmarks, such as AVeriTeC~\cite{averitec}, MOCHEG~\cite{MOCHEG}, VERITE~\cite{VERITE}, and AVerImaTeC~\cite{AVerImaTeC}. 
As discussed in Sec.~\ref{sec:intro}, these benchmarks risk contamination, which can compromise the reliability of evaluation results. 
To address this, the emerging dynamic evaluation paradigm continuously introduces unseen, time-stamped data to enhance robustness and mitigate such contamination risks~\cite{livebench, livecodebench, swelive}. 
For example, LiveBench~\cite{livebench} automatically sources new questions across various domains, such as reasoning, coding, math, and writing, from recent public benchmarks and updates its evaluation data on a monthly basis. 
For MAFC, two dynamic benchmarks exist: XFACTA~\cite{xfacta}, which is no longer actively maintained (last updated in August 2025), and VERITAS~\cite{veritas}, the SOTA benchmark that aggregates real-world claims from 108 professional fact-checking agencies and updates quarterly through a fully automated pipeline.
\textbf{Nonetheless, there is still limited understanding of the extent of contamination in existing static and dynamic MAFC benchmarks and its impact on evaluation outcomes.} 
In this work, we leverage the evidence sufficiency evaluation pipeline from AVeriTeC for detecting potential contamination. 
Our analysis (Sec.~\ref{sec:exp}) reveals that dynamic MAFC benchmarks remain susceptible to contamination, which can significantly distort evaluation results. 
To mitigate this, we filter out potentially contaminated samples to provide a contamination-controlled assessment of SOTA LLMs.

Closest to our work, \citet{yoon2025hypothetical} examine contamination in MAFC with systems using LLM-based query expansion. However, their study is limited to systems reliant on human-specified workflows and only evaluates existing static benchmarks. In contrast, we study SOTA MAFC systems that operate as fully autonomous agents and analyze emerging dynamic benchmarks, providing a more forward-looking assessment of contamination risks in MAFC.

\begin{figure*}[t]
\centering 
\includegraphics[width=\linewidth]{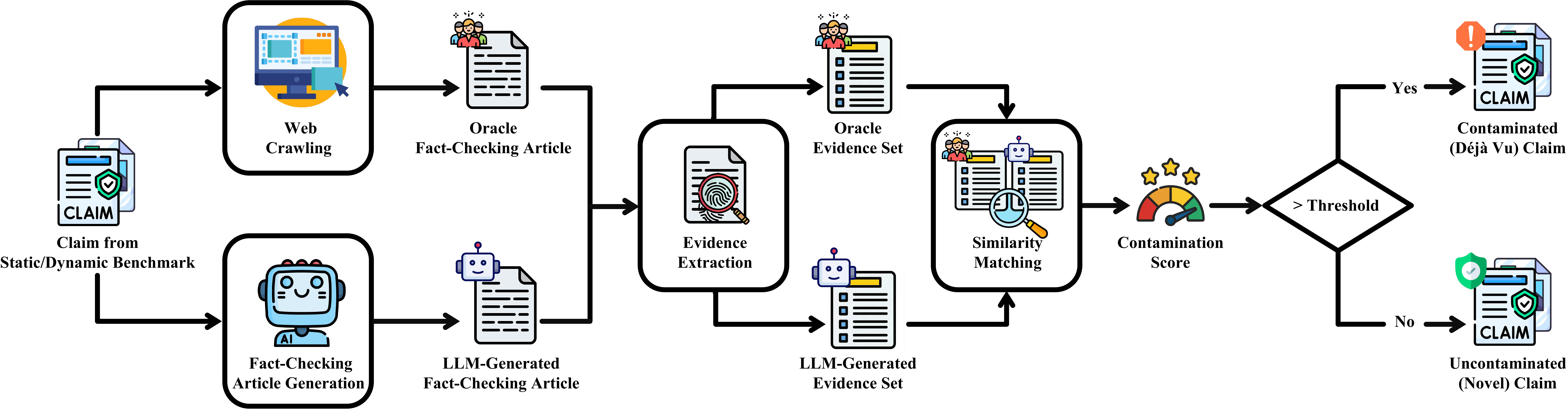}
\caption{An overview of contamination detection via an evidence sufficiency evaluation pipeline.}
\label{fig:pipeline}
\end{figure*}

\section{Contamination Detection for MAFC}\label{sec:method}
To address the RQs motivating this work, we first define contamination for the MAFC task, then introduce our contamination detection pipeline, and finally describe the benchmarks we use.\footnote{Example prompts used in this section are provided in Appendix~\ref{app:prompt}.}

\subsection{Definition of Contamination in MAFC}
MAFC systems predict the veracity label of a claim \(c\). 
This task unfolds in two sequential stages: (1) \textbf{evidence retrieval}, where the system gathers a set of multimodal evidence items \(\mathcal{E} = \{e_1, \dots, e_k\}\) from external sources (e.g., the open web), each item potentially containing textual and/or visual information, to support or refute \(c\); and (2) \textbf{claim verification}, where the system aggregates and reasons over \(\mathcal{E}\) to assign a veracity label to \(c\).
Prior studies~\cite{averitec_challenge, evidence_all_you_need, he2025debating} demonstrate that MAFC performance is largely determined by the evidence retrieval stage. 
In light of this,
we consider a claim as \emph{contaminated} for a target LLM when the LLM possesses sufficient parametric knowledge matching the evidence used by human fact-checkers.
While parametric knowledge can help verify claims about past events, MAFC systems are primarily intended for emerging events with little or no public evidence~\cite{veritas, xfacta}. 
Thus, this definition captures whether the model can bypass the core MAFC challenge of retrieving evidence from external, in-the-wild sources.

\subsection{Contamination Detection Pipeline}
Based on the definition, the next question becomes: \emph{how can we evaluate the sufficiency of the evidence in LLMs' internal knowledge with respect to the oracle evidence used by humans?}
To address this, we adopt the established evidence sufficiency evaluation pipeline proposed in AVeriTeC~\cite{averitec} for potential contamination detection. 
This choice is aligned with the definition: 
we measure contamination risk by the extent to which the LLMs can generate sufficient oracle evidence from their parametric knowledge.
As shown in Fig.~\ref{fig:pipeline}, this pipeline comprises the following steps.

\textbf{Step 1: Fact-checking Article Generation.}
Given a claim $c$ in a MAFC benchmark, we prompt a target LLM $m$ to generate a fact-checking article $\tilde{a}$ for~$c$.
This setup is intended to elicit the internal parametric knowledge available to the model.

\textbf{Step 2: Evidence Extraction.}
After obtaining the LLM-generated article $\tilde{a}$ for claim $c$ and its corresponding human-written oracle fact-checking article $a$, crawled from fact-checking agencies, we employ GPT-4o-Mini to extract evidence items from both the oracle article \(a\) and the LLM-generated article \(\tilde{a}\), yielding the sets \(\mathcal{E} = \{e_1, \ldots, e_n\}\) and \(\tilde{\mathcal{E}} = \{\tilde{e}_1, \ldots, \tilde{e}_{n'}\}\), respectively.\footnote{Appendix~\ref{app:robust} validates that evidence extraction is robust to different LLMs.}
Each evidence item is a claim-relevant textual statement extracted from a source article. 
Although textual in form, it may encode interpretations or reasoning derived from multimodal content (e.g., images, audio, or video).
This extraction step reduces stylistic mismatches between human-written and LLM-generated articles while preserving the core content and removing extraneous elements such as cookie notices, advertisements, and other UI noise. Applying the same extraction to both articles ensures a fair comparison.

\textbf{Step 3: Similarity Matching.}
After extracting the evidence, we require a method to assess the similarity between $\mathcal{E}$ (oracle) and $\tilde{\mathcal{E}}$ (LLM-generated). 
Here, we follow the Hungarian-matching protocol from AVeriTeC~\cite{averitec}, where \(\mathcal{E}\) plays the role of the human-annotated reference evidence, while \(\tilde{\mathcal{E}}\) serves as the ``retrieved'' evidence derived purely from the target LLM's internal knowledge. Formally, for a generated evidence set \(\tilde{\mathcal{E}}\) and an oracle evidence set \(\mathcal{E}\), we compute the claim-level contamination score as follows:
\begin{equation}
s(\tilde{\mathcal{E}}, \mathcal{E}) = \frac{1}{|\mathcal{E}|}\max\sum_{\tilde{e}_i \in \tilde{\mathcal{E}}}\sum_{e_j \in \mathcal{E}} f(\tilde{e}_i, e_j) X(\tilde{e}_i, e_j),
\end{equation}
where \(f\) is a pairwise similarity function and \(X(\tilde{e}_i,e_j) \in \{0,1\}\) encodes a one-to-one assignment obtained via the Hungarian algorithm. Normalization by \(|\mathcal{E}|\) prevents artificial inflation from over-generation while penalizing omissions of oracle evidence. The resulting score quantifies the degree of similarity between generated evidence and oracle evidence.
Using this pipeline, we compute a claim-level contamination score $s(\tilde{\mathcal{E}}, \mathcal{E})$ for each claim $c$ with respect to a target LLM $m$. 
A claim $c$ is flagged as potentially contaminated for $m$ if its score satisfies \(s \ge \tau\), where \(\tau\) is a predefined threshold. 
To obtain a benchmark-level contamination score $S$ for LLM \(m\), we average the claim-level scores across all claims:  
\begin{equation} 
S = \frac{1}{|\mathcal{C}|} \sum_{c \in \mathcal{C}} s(\tilde{\mathcal{E}}_c, \mathcal{E}_c),
\end{equation}
where \(\mathcal{C}\) denotes the set of claims in the benchmark.

\subsection{Evaluated MAFC Benchmarks}
To apply this pipeline in practice, we select two complementary MAFC benchmarks: one established static benchmark and one newly constructed dynamic benchmark.

\paragraph{Static Benchmark.} 
To evaluate the contamination level of static MAFC benchmarks, 
we use the SOTA AVeriTeC~\cite{averitec} benchmark, which consists of claims from 50 fact-checking agencies, each annotated by professional fact-checkers. 
This benchmark addresses key limitations of previous MAFC benchmarks, 
such as temporal evidence leakage and evidence insufficiency, 
and is widely adopted by SOTA systems~\cite{defame, he2025debating, hero}. 
For our experiments, we use the AVeriTeC development set, which contains 500 claims. The most recent claim in this benchmark is dated October 31, 2020, posing a significant risk of contamination, as the knowledge cut-off dates for SOTA LLMs are typically no earlier than 2023 (see Table~\ref{tab:llm}). 
To ensure data quality, we first deduplicate the claims, resulting in 491 unique claims, then crawl their corresponding oracle fact-checking articles using the source URLs provided in AVeriTeC.

\paragraph{Dynamic Benchmark.}
To empirically evaluate post-cut-off claims, we adopt the data collection protocol of VERITAS and ClaimReview2024+~\cite{veritas, defame} to source up-to-date fact-checks from the ClaimReview project, which aggregates structured articles from fact-checking agencies worldwide.\footnote{As XFACTA is no longer maintained and VERITAS is not open-source at the time of writing, we curate our own dynamic benchmark to support this evaluation using a similar claim source and curation pipeline.
Please note that our main contribution is the empirical analysis of contamination, rather than introducing a new benchmark. 
The curated benchmark is solely for enabling our analysis. 
}
Given that the knowledge cut-off date of the SOTA LLM GPT-5.2 is August 2025, we collect only claims published in Q4 2025 (Oct.--Dec. 2025) and denote the resulting benchmark as ClaimReview2025Q4.\footnote{We adapt collection code from MisinfoMe~\cite{misinfome} and CimpleKG~\cite{cimplekg}.} 
To maintain comparability with AVeriTeC, we restrict our collection to English claims.
For source credibility, we retain only claims fact-checked by the agencies that are signatories of the International Fact-Checking Network (IFCN).\footnote{Full list available at \url{https://ifcncodeofprinciples.poynter.org/signatories}.}
Then, we deduplicate repeated claims from different agencies.
Finally, the raw claim text provided by ClaimReview may be unsuitable for MAFC evaluation. 
It may reveal the verdict (e.g., ``a fake video...'', ``a manipulated image...''), 
consist of incomplete sentences, 
or contain unnecessary meta-information such as ``A viral social media post claims...'', instead of directly expressing a factual proposition.
Thus, we leverage GPT-4o-Mini to filter out these malformed claims.
After this processing, 
the resulting ClaimReview2025Q4 benchmark contains 901 precise, self-contained claims.
As with AVeriTeC, we crawl the oracle fact-checking articles for each claim from their source URLs.

\section{Experiments}\label{sec:exp}
This section details our empirical evaluation, designed to address the four research questions (RQs) posed in Sec.~\ref{sec:intro}.

\subsection{RQ1: The Extent of Contamination}\label{sec:rq1}

To address RQ1, we apply the evidence sufficiency evaluation pipeline to evaluate the contamination levels in existing benchmarks.

\paragraph{Experimental Setup.}
As described in Sec.~\ref{sec:method}, we quantify contamination by measuring the pairwise similarity scores between evidence items in LLM-generated fact-checking articles and those in human-written oracle articles for each claim, on both static and dynamic benchmarks (i.e., AVeriTeC and ClaimReview2025Q4).

To compute the pairwise scores, we consider both \textbf{textual} and \textbf{semantic} similarities. 
For textual similarity, we follow the default configuration in AVeriTeC~\cite{averitec}, using \emph{METEOR}\footnote{\url{https://www.nltk.org/_modules/nltk/translate/meteor_score.html}} as the pairwise similarity function $f$ and setting $\tau = 25\%$.
For semantic similarity, 
we employ two SOTA text representation models: \emph{Gemma-Emb-0.3B}\footnote{\url{https://huggingface.co/google/embeddinggemma-300m}} 
and 
\emph{Qwen3-Emb-0.6B}\footnote{\url{https://huggingface.co/Qwen/Qwen3-Embedding-0.6B}} 
and instantiate $f$ as the cosine similarity between the embeddings of each pair of evidence items. 
To determine the contamination threshold for our semantic metrics, we randomly sampled 180 claims (20\% of ClaimReview2025Q4) for human evaluation. 
Two annotators with postgraduate-level expertise in journalism and/or fact-checking, both proficient in English, independently assessed each claim for contamination. 
They agreed on 97.8\% of the claims; the remaining disagreements were resolved through discussion. 
We used these adjudicated labels as ground truth and performed a threshold-sweep analysis over $\tau \in [0,1]$.\footnote{Please see Appendix~\ref{app:threshold_sweep} for the complete threshold-sweep analysis over $\tau \in [0,1]$.}
At approximately $\tau = 0.6$, contamination labels exhibited the highest agreement with human annotations: 75.6\% for Gemma-Emb-0.3B and 82.2\% for Qwen3-Emb-0.6B. Thus, we selected $\tau = 0.6$ as an empirically grounded contamination threshold for semantic metrics.

To benchmark contamination levels in SOTA LLMs, we evaluate both \textbf{proprietary} models, including OpenAI's \textit{GPT-5.2}, \textit{GPT-4o-Mini}, Google's \textit{Gemini-3.0-Pro}, and \textit{Gemini-3.0-Flash}, and two \textbf{open-source} models (DeepSeek's \textit{DeepSeek-V3.2} and Alibaba's \textit{Qwen3.5-122B-A10B}). Table~\ref{tab:llm} provides an overview of these LLMs.
All models use a temperature of 0.01 to encourage deterministic behavior. 

\begin{table}[h]
\centering
\caption{Overview of the evaluated LLMs.}
\label{tab:llm}
\small
\resizebox{0.85\linewidth}{!}{%
\begin{tabular}{l c c}
\toprule
\textbf{LLMs} & \textbf{Open-Source} & \textbf{Knowledge Cut-Off} \\
\midrule
GPT-5.2 & $\times$  & Aug. 2025 \\
GPT-4o-Mini & $\times$  & Oct. 2023 \\
Gemini-3.0-Pro & $\times$  & Jan. 2025 \\
Gemini-3.0-Flash & $\times$  & Jan. 2025 \\
DeepSeek-V3.2 & $\checkmark$ & Unknown \\
Qwen3.5-122B-A10B & $\checkmark$ & Unknown \\
\bottomrule
\end{tabular}
}
\end{table}

\paragraph{Results.}

\begin{table*}[htbp]
\centering
\caption{Average contamination scores between LLM-generated fact-checking articles and oracle articles from two benchmarks for six models across three metrics. The highest score in each column is \textbf{bolded}; the second-highest is \underline{underlined}.}
\label{tab:sim}
\resizebox{\textwidth}{!}{%
\begin{tabular}{lcccccc}
\toprule
\multirow{2}{*}{\textbf{LLMs}} 
& \multicolumn{2}{c}{\textbf{METEOR}}
& \multicolumn{2}{c}{\textbf{Gemma-Emb-0.3B}}
& \multicolumn{2}{c}{\textbf{Qwen3-Emb-0.6B}} \\
\cmidrule(lr){2-3} \cmidrule(lr){4-5} \cmidrule(lr){6-7}
& AVeriTeC & ClaimReview2025Q4
& AVeriTeC & ClaimReview2025Q4
& AVeriTeC & ClaimReview2025Q4 \\
\midrule
GPT-5.2                 & 23.78\% & 19.15\% & 65.28\% & 54.12\% & 63.82\% & 53.47\% \\
GPT-4o-Mini             & \textbf{26.80\%} & \underline{21.93\%} & 63.70\% & 55.28\% & 61.67\% & 53.11\% \\
Gemini-3.0-Pro    & \underline{26.58\%} & 21.63\% & \textbf{67.00\%} & \underline{57.82\%} & \textbf{65.78\%} & \underline{56.89\%} \\
Gemini-3.0-Flash & 25.42\% & 20.90\% & 65.30\% & 57.12\% & 64.00\% & 55.72\% \\
DeepSeek-V3.2         & 24.65\% & 20.45\% & 64.88\% & 57.05\% & 63.59\% & 55.65\% \\
Qwen3.5-122B-A10B      & 26.08\% & \textbf{23.16\%} & \underline{66.58\%} & \textbf{60.38\%} & \underline{64.87\%} & \textbf{58.48\%} \\
\bottomrule
\end{tabular}
}
\end{table*}

Table~\ref{tab:sim} shows the average textual and semantic contamination scores between LLM-generated fact-checking articles and oracle articles, calculated using three metrics across six LLMs. 
Results are averaged over at least three repeated trials.
The results reveal several key findings.
\textbf{Finding 1.1: The dynamic benchmark can effectively alleviate contamination.}
The dynamic benchmark (ClaimReview2025Q4) consistently exhibits lower contamination levels than the static benchmark (AVeriTeC) across all six models.
Under METEOR, the contamination scores decrease by 
2.92--4.95 percentage points; under the two semantic metrics, the reductions are 6.20--11.16 percentage points.
\textbf{Finding 1.2: Strong contamination signals are widespread across LLMs on AVeriTeC; on ClaimReview2025Q4, peak scores are concentrated in particular models.} 
On AVeriTeC, nearly all models exhibit average contamination scores above the contamination thresholds, with only two exceptions under the strict METEOR metric, which requires exact textual matches. 
GPT-4o-Mini shows the strongest contamination level under METEOR (26.80\%), while Gemini-3.0-Pro achieves the highest scores under both semantic metrics (67.00\% on Gemma-Emb-0.3B and 65.78\% on Qwen3-Emb-0.6B). 
In contrast, on ClaimReview2025Q4, Qwen3.5-122B-A10B attains the highest scores across all three metrics: 23.16\% (METEOR), 60.38\% (Gemma-Emb-0.3B), and 58.48\% (Qwen3-Emb-0.6B).

\begin{table}[htbp]
\centering
\caption{Proportion of potentially contaminated claims in ClaimReview2025Q4 for each LLM under different metrics. The intersection column indicates claims identified as contaminated by all three metrics.}
\label{tab:proportion}
\resizebox{\columnwidth}{!}{%
\begin{tabular}{lcccc}
\toprule
\textbf{LLMs} & \textbf{METEOR} & \textbf{Gemma-Emb-0.3B} & \textbf{Qwen3-Emb-0.6B} & \textbf{Intersection} \\
\midrule
GPT-5.2 & 22.31\% & 38.62\% & 38.51\% & 17.09\%  \\
GPT-4o-Mini  & \underline{32.52\%} & 42.62\% & 36.40\% & \underline{22.97\%}\\
Gemini-3.0-Pro & 28.41\% & \underline{49.06\%} & \underline{46.95\%} & 22.86\% \\
Gemini-3.0-Flash & 25.97\% & 44.51\% & 40.73\% & 20.09\% \\
DeepSeek-V3.2 & 25.19\% & 45.95\% & 41.62\% & 20.31\% \\
Qwen3.5-122B-A10B & \textbf{35.18\%} & \textbf{55.94\%} & \textbf{50.94\%} & \textbf{29.30\%} \\

\bottomrule
\end{tabular}
}
\end{table}

\begin{figure}[htbp]
\centering

\begin{subfigure}{\linewidth}
    \centering
    \includegraphics[width=0.88\linewidth]{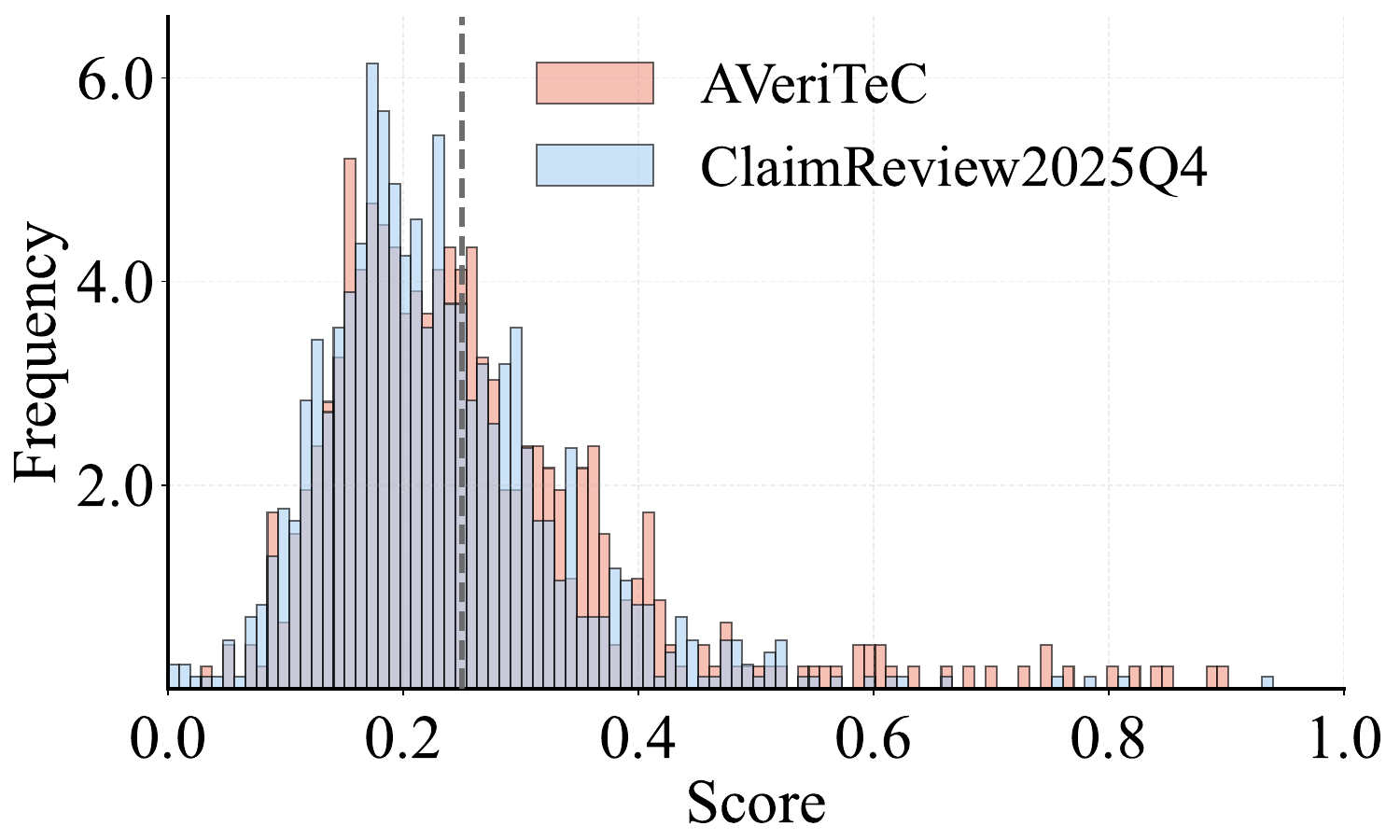}
    \caption{METEOR}
    \label{fig:rq1-qwen-meteor}
\end{subfigure}

\begin{subfigure}{\linewidth}
    \centering
    \includegraphics[width=0.88\linewidth]{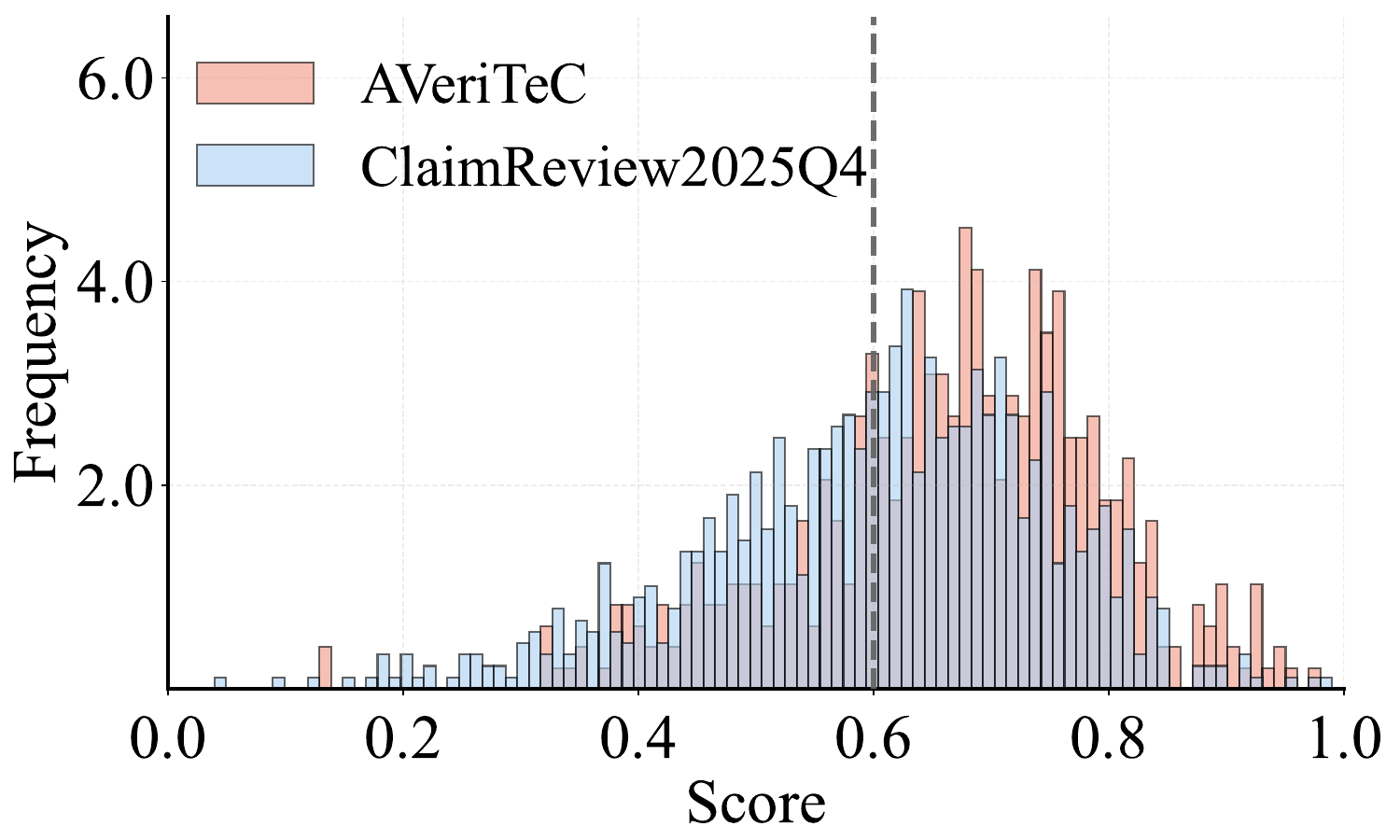}
    \caption{Gemma-Emb-0.3B}
    \label{fig:rq1-qwen-gte}
\end{subfigure}

\begin{subfigure}{\linewidth}
    \centering
    \includegraphics[width=0.88\linewidth]{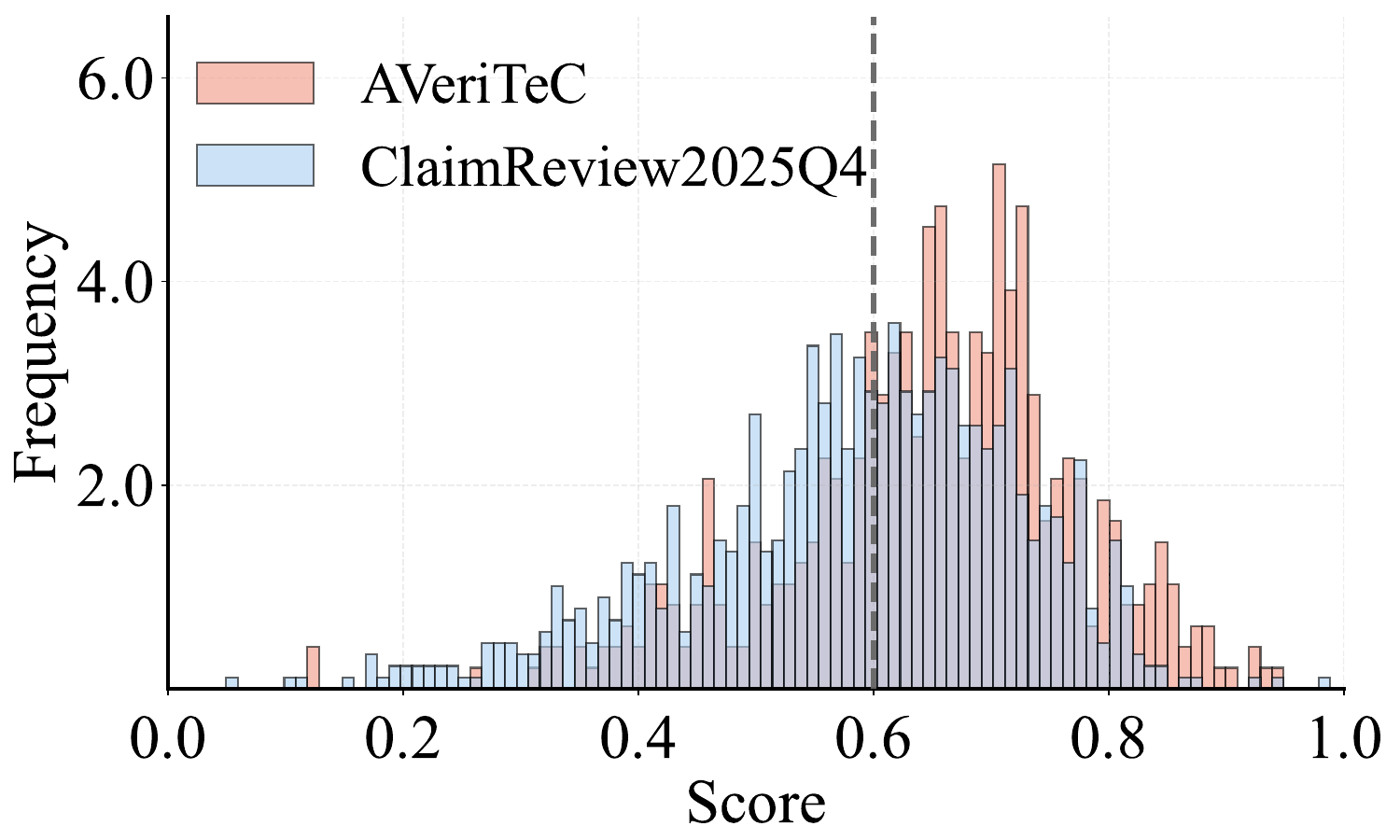}
    \caption{Qwen3-Emb-0.6B}
    \label{fig:rq1-qwen-qwen3}
\end{subfigure}

\caption{Contamination score distributions of Qwen3.5-122B-A10B on AVeriTeC and ClaimReview2025Q4.}
\label{fig:rq1-qwen-distribution}
\end{figure}

Table~\ref{tab:proportion} further reports the proportion of contaminated claims in ClaimReview2025Q4 identified by the criterion $s(\tilde{\mathcal{E}}, \mathcal{E}) \ge \tau$ for each metric and each model. 
The results provide a more fine-grained view of contamination in the dynamic benchmark, yielding the following findings.
\textbf{Finding 1.3: A non-trivial portion of claims in the dynamic benchmark still faces contamination risk across all models.}
When considering the intersection of the three metrics (i.e., exceeding the thresholds for all three), the proportion of potentially contaminated claims ranges from 17.09\% to 29.30\%, depending on the model.
This indicates that a dynamic evaluation protocol substantially reduces, but does not fully eliminate, the risk of contamination.
\textbf{Finding 1.4: Semantic metrics identify a markedly larger contaminated subset than lexical matching.}
For every model, the proportions under Gemma-Emb-0.3B and Qwen3-Emb-0.6B are consistently much higher than those under METEOR, indicating that contamination often appears in semantically similar reformulations rather than near-verbatim reuse alone.
This finding further suggests that relying only on lexical overlap can underestimate the practical contamination risk in MAFC benchmarks.
\textbf{Finding 1.5: Qwen3.5-122B-A10B is the most contaminated model on the dynamic benchmark.}
It has the highest contaminated-claim ratio across all three metrics: 35.18\% (METEOR), 55.94\% (Gemma-Emb-0.3B), and 50.94\% (Qwen3-Emb-0.6B).
Notably, since the knowledge cut-off date for Qwen3.5-122B-A10B is unknown (see Table~\ref{tab:llm}), the results may indicate that the model could include online content as recent as Q4 2025. 
To further examine the extent of contamination, Fig.~\ref{fig:rq1-qwen-distribution} visualizes the claim-level contamination score distribution for Qwen3.5-122B-A10B on AVeriTeC and ClaimReview2025Q4. 
This visualization reveals \textbf{Finding 1.6: A subset of claims in the dynamic benchmark exhibits extremely high contamination risk.}
All three distributions retain a visible right tail above the contamination thresholds, with some samples showing exceptionally high scores (e.g., over 80\%).

\subsection{RQ2: The Cause of Contamination}\label{sec:rq2}
Given that a subset of claims in the dynamic benchmark exhibits strong contamination despite their source articles being published after the LLMs' knowledge cut-off dates, we investigate the underlying reasons for this unexpected contamination.

\paragraph{Experimental Setup.}
We present a qualitative case study of four claims from ClaimReview2025Q4, all published in \textbf{Q4 2025}. Despite their recent publication, these claims exceed contamination thresholds for all three similarity metrics across all six models.

\paragraph{Results.}
The four representative examples presented in Table~\ref{tab:rq2_cases} highlight two main sources of contamination in the dynamic benchmark, summarized as follows.
\textbf{Finding 2.1: Contamination arises from claims that directly reference facts available before the LLMs' knowledge cut-off dates.}  
Even when fact-checking articles are published post-cut-off, claims about historical events, established policies, or other pre-existing facts can often be verified using information already present in the LLM's training data.
For example, \emph{claim 1} centers on EMTALA, a law enacted in \textbf{1986} that is widely documented in sources like Wikipedia.\footnote{\url{https://en.wikipedia.org/wiki/Emergency_Medical_Treatment_and_Active_Labor_Act}} Similarly, \emph{claim 2} concerns Scarborough's canceled fireworks display in \textbf{2022}, as widely reported by outlets such as the BBC,\footnote{\url{https://www.bbc.com/news/uk-england-york-north-yorkshire-64139048}} and \emph{claim 3} discusses the free TV licence policy, established in \textbf{2020} and repeatedly covered thereafter.\footnote{\url{https://www.gov.uk/free-discount-tv-licence}}
As a result, direct or contextual knowledge of these facts was likely incorporated into the pretraining data of LLMs, bypassing the need for external evidence retrieval that MAFC systems are designed to perform.
\textbf{Finding 2.2: Contamination can also result from synthesizing multiple pre-cut-off facts.}  
Some claims may be novel in their formulation (never having appeared previously), but their veracity can be determined by synthesizing multiple pieces of knowledge available before the cut-off. For instance, \emph{claim 4} involves Barron Trump's eligibility for the U.S. Senate in 2028. While this specific claim is new, it can be refuted by combining two well-established facts: Barron Trump's birth year (\textbf{2006}) and the age requirement for senators (30 years, set in \textbf{1787}).

This case study shows that many contamination signals stem from claims that, although newly published, are verifiable using public information available before the cut-off.

\begin{table*}[htbp]
\caption{Contaminated claims from ClaimReview2025Q4. Content is condensed for brevity while preserving original meaning. Contaminated oracle fact-check snippets are highlighted in \flaw{purple}; matched LLM-generated snippets are shown in \good{green}.}
\centering
\footnotesize
\resizebox{0.96\textwidth}{!}{
\begin{tabular}{p{4cm} p{4.9cm} p{4.9cm}}
\toprule
\textbf{Claim} & \textbf{Oracle Fact-Checks} & \textbf{LLM-Generated Fact-Checks}  \\ \midrule
Former U.S. President Ronald Reagan signed the Emergency Medical Treatment and Active Labor Act, which required emergency rooms to provide emergency health care to anyone, regardless of immigration status, in 1986. \textbf{(Claim 1, published on Nov. 1, 2025)}
& \flaw{Reagan did sign the Emergency Medical Treatment and Active Labor Act in 1986.} The act required ``participating hospitals'' that offer emergency services to \flaw{provide care regardless of patients' ability to pay or immigration status.}
& \good{The Emergency Medical Treatment and Active Labor Act (EMTALA) was signed into law by President Ronald Reagan on April 7, 1986.} The enforcement and legal interpretation of the law mandate that emergency rooms provide screening and stabilization \good{to all individuals, including those regardless of their immigration status.}
\\ \midrule
A town in England canceled a New Year's Eve fireworks show so a walrus could sleep. 
\textbf{(Claim 2, published on Dec. 25, 2025)}
& A real BBC News article confirmed the basics of the story as shared on social media: \flaw{In 2022, the English town of Scarborough canceled a New Year's Eve fireworks} display at the last minute over fears \flaw{it could cause distress to a walrus.}
& \good{In December 2022}, the seaside \good{town of Scarborough} in North Yorkshire, England, officially \good{canceled its New Year's Eve fireworks} display to \good{protect the welfare of a visiting Arctic walrus.}
 \\ \midrule
UK residents over the age of 60 can claim a free lifetime exemption from paying the TV licence.
\textbf{(Claim 3, published on Nov. 28, 2025)}
& From 2020, the BBC decided to pay for free TV licences for the households of \flaw{anyone over the age of 75 receiving Pension Credit.} There is no such blanket lifetime exemption for the over 60s.
& In the United Kingdom, \good{the age threshold for a free TV licence is 75, not 60.} Further, eligibility for those aged 75 and over is not automatic; it is contingent \good{upon receiving specific state benefits.}
\\ \midrule
Barron Trump announced a 2028 Senate run. 
\textbf{(Claim 4, published on Nov. 22, 2025)}
& 
The \flaw{U.S. Constitution requires senators to be at least 30 years old}. \flaw{Barron Trump was born on March 20, 2006}, making him 19 years old. So in 2028 he would still not meet the constitutional minimum age of 30 for qualification to serve in the U.S. Senate. 
& 
\good{Barron Trump was born on March 20, 2006.} By the end of 2028, he would be 22 years old. He will not turn 30 until 2036; therefore, he is constitutionally ineligible to run for or serve in the \good{U.S. Senate} during the 2028 election cycle because the \good{constitutional minimum age requirement is 30 years}.
\\ 
\bottomrule
\end{tabular}
}
\label{tab:rq2_cases}
\end{table*}

\begin{table}[t]
\centering
\setlength{\tabcolsep}{10pt}

\caption{MAFC performance on the contaminated and uncontaminated subsets of ClaimReview2025Q4. Red superscripts indicate drops in performance when moving from the contaminated to the uncontaminated subset; $*$ marks statistically significant drops ($p<0.05$, bootstrap test).}
\label{tab:contaminated}

\resizebox{\columnwidth}{!}{%
\begin{tabular}{l cc cc}
\toprule

\multirow{2}{*}{\textbf{LLMs}} 
& \multicolumn{2}{c}{\textbf{Contaminated}}
& \multicolumn{2}{c}{\textbf{Uncontaminated}} \\
\cmidrule(lr){2-3} \cmidrule(lr){4-5}

& Accuracy & Macro-F1 
& Accuracy & Macro-F1 \\
\midrule

GPT-5.2                
& 61.04\% 
& 57.68\% 
& 52.88\%\rlap{\textsuperscript{\textcolor{red}{\scriptsize -8.16\%}}}
& \textbf{52.81\%}\rlap{\textsuperscript{\textcolor{red}{\scriptsize -4.87\%}}}
\\

GPT-4o-Mini            
& 65.22\% 
& 56.41\% 
& 55.62\%\rlap{\textsuperscript{\textcolor{red}{\scriptsize -9.60\%$^*$}}}
& 46.82\%\rlap{\textsuperscript{\textcolor{red}{\scriptsize -9.59\%$^*$}}}
\\

Gemini-3.0-Pro   
& 71.36\% 
& \underline{61.07\%} 
& 60.86\%\rlap{\textsuperscript{\textcolor{red}{\scriptsize -10.50\%$^*$}}}
& 52.50\%\rlap{\textsuperscript{\textcolor{red}{\scriptsize -8.57\%$^*$}}}
\\

Gemini-3.0-Flash 
& \underline{74.59\%} 
& \textbf{61.22\%} 
& \textbf{66.81\%}\rlap{\textsuperscript{\textcolor{red}{\scriptsize -7.78\%$^*$}}}
& 49.88\%\rlap{\textsuperscript{\textcolor{red}{\scriptsize -11.34\%$^*$}}}
\\

DeepSeek-V3.2          
& 57.92\% 
& 54.79\% 
& 52.92\%\rlap{\textsuperscript{\textcolor{red}{\scriptsize -5.00\%}}}
& \underline{52.57\%}\rlap{\textsuperscript{\textcolor{red}{\scriptsize -2.22\%}}}
\\

Qwen3.5-122B-A10B      
& \textbf{74.62\%} 
& 57.48\% 
& \underline{63.74\%}\rlap{\textsuperscript{\textcolor{red}{\scriptsize -10.88\%$^*$}}}
& 47.16\%\rlap{\textsuperscript{\textcolor{red}{\scriptsize -10.31\%$^*$}}}
\\

\bottomrule
\end{tabular}%
}

\end{table}

\subsection{RQ3: The Impact of Contamination}\label{sec:rq3}
Given that MAFC benchmarks risk contamination, we evaluate how such contamination affects MAFC performance.

\paragraph{Experimental Setup.}
We instantiate the SOTA MAFC framework DEFAME~\cite{defame}, using each LLM selected in RQ1 as the backbone.
DEFAME employs the standard ReAct-style~\cite{react} agentic paradigm, iteratively reasoning and selecting actions, making it a suitable default for our evaluation.
In our experiments, LLMs within DEFAME access the open web by invoking Google Web Search and/or Image Search tools via the Serper API\footnote{\url{https://serper.dev}}, which returns the top three most relevant web pages per query, including text, images, diagrams, etc.
For each LLM, we define a \textbf{model-specific contaminated subset} by taking the intersection of claims that exceed contamination thresholds across all three similarity metrics.
The remaining claims constitute the \textbf{uncontaminated subset}.
Unlike prior benchmarks that group claims by publication date and thus confound contamination with temporal distribution shifts, both subsets are drawn from the same sources and time frame.\footnote{Appendix~\ref{app:disentangle} validates that they are highly comparable in topical and stylistic features.}
We then compare \textbf{Accuracy} and \textbf{Macro-F1} between these two subsets.

\paragraph{Results.}
Table~\ref{tab:contaminated} compares MAFC performance on the contaminated and uncontaminated subsets for each model.
The results reveal four key findings.
\textbf{Finding 3.1: Contamination can significantly inflate evaluation results.}
Both Accuracy and Macro-F1 decrease for every model when moving from the contaminated subset to the uncontaminated subset, with statistically significant drops observed for Qwen3.5-122B-A10B, Gemini-3.0-Pro, Gemini-3.0-Flash, and GPT-4o-Mini.
\textbf{Finding 3.2: The magnitude of this inflation varies substantially across models.}
Qwen3.5-122B-A10B exhibits the largest Accuracy drop, decreasing by 10.88 percentage points, and also shows a substantial Macro-F1 decline of 10.31 percentage points. Gemini-3.0-Pro, Gemini-3.0-Flash, and GPT-4o-Mini also exhibit substantial declines, with Accuracy/Macro-F1 drops of (10.50/8.57), (7.78/11.34), and (9.60/9.59) percentage points, respectively, while GPT-5.2 and DeepSeek-V3.2 show no statistically significant degradation.
\textbf{Finding 3.3: The most contaminated model also suffers the largest Accuracy decline.}
In Table~\ref{tab:sim}, Qwen3.5-122B-A10B exhibits the strongest contamination levels on ClaimReview2025Q4 across all three metrics. 
Consistently, this model also demonstrates the largest Accuracy drop when comparing contaminated to uncontaminated claims.
This alignment is consistent with a positive association between the extent of contamination and the degree of benchmark score inflation, 
which also provides converging evidence that our contamination detection pipeline captures a meaningful evaluation artifact.
\textbf{Finding 3.4: Contamination can distort MAFC performance rankings of LLMs.} 
On the contaminated subset, Gemini-3.0-Flash achieves the highest Macro-F1 score (61.22\%);
however, on the uncontaminated subset, the best Macro-F1 shifts to GPT-5.2 (52.81\%).
This indicates that benchmark contamination can obscure models' relative capabilities when fact-checking genuinely unseen claims, motivating a contamination-controlled evaluation.

\begin{table}[h]
\centering
\small
\caption{Trajectory-level reformulation statistics on contaminated (Con.) and uncontaminated (Uncon.) subsets. Values report the average number of transitions per claim. $*$ indicates a statistically significant increase relative to the corresponding Con. counterpart ($p<0.05$; bootstrap test).}
\label{tab:why}
\resizebox{\columnwidth}{!}{%
\begin{tabular}{llcccc}
\toprule
\textbf{Model} & \textbf{Type} & \textbf{Spe.} & \textbf{Gen.} & \textbf{Exp.} & \textbf{Rep.} \\
\midrule
\multirow{2}{*}{Qwen3.5-122B-A10B}
    & Con. 
    & 0.53 
    & \textbf{0.47}
    & 0.89 
    & \textbf{0.04} \\
    & Uncon. 
    & \textbf{0.60} 
    & 0.45 
    & \textbf{1.24}$^*$ 
    & 0.03 \\
\midrule
\multirow{2}{*}{Gemini-3.0-Pro}
    & Con. 
    & 0.80 
    & 0.49 
    & 1.22 
    & 0.11 \\
    & Uncon. 
    & \textbf{0.94} 
    & \textbf{0.56} 
    & \textbf{1.54}$^*$ 
    & \textbf{0.12} \\
\bottomrule
\end{tabular}
}
\end{table}

\paragraph{Analysis.}
To further understand \textit{why} contamination changes downstream MAFC performance, 
we follow prior work~\cite{query1, query2, ning2026agentic} and analyze \textbf{trajectory-level reformulations} during the fact-checking process.
Specifically, for each fact-checking trajectory, we follow \citet{ning2026agentic} and use GPT-5-Nano (temperature 0.01, with three repeated trials) to label every adjacent search query pair $(q_k \rightarrow q_{k+1})$ as one of four types: 
\textbf{Specialization (Spe.)} (narrowing the query by adding constraints), 
\textbf{Generalization (Gen.)} (broadening the query by relaxing constraints), 
\textbf{Exploration (Exp.)} (pivoting to a different facet within the same topic), or 
\textbf{Repetition (Rep.)} (issuing an identical or near-duplicate reformulation).\footnote{\citet{ning2026agentic} provide examples of these categories and validate the robustness of their method to different LLMs. We follow their default setup to use GPT-5-Nano.}
We then compute the average number of each reformulation type for both contaminated and uncontaminated subsets, focusing on the top two models with the largest Accuracy declines as shown in Table~\ref{tab:contaminated}, i.e., Qwen3.5-122B-A10B and Gemini-3.0-Pro.
As shown in Table~\ref{tab:why}, the average counts of Specialization, Generalization, and Repetition remain statistically indistinguishable across contaminated and uncontaminated subsets. However, contaminated trajectories consistently devote significantly fewer steps to Exploration, which suggests \textbf{Finding 3.5: Contamination enables models to retrieve relevant evidence more precisely, bypassing comprehensive exploration of the evidence space.} When facing contaminated claims, LLMs retrieve relevant evidence efficiently, without thoroughly considering alternative queries or being distracted by irrelevant details. As a result, contamination may inflate accuracy scores by shortcutting evidence retrieval.

\subsection{RQ4: Contamination-Controlled Evaluation}\label{sec:clean_eval}
Motivated by the above findings, which highlight the significant impact of contamination on evaluation results, we further assess the LLMs in a stricter contamination-controlled setting.

\paragraph{Experimental Setup.}
Using the same DEFAME framework and six LLM backbones, we obtain a unified contamination-controlled evaluation set by retaining only claims that are uncontaminated for all six models under METEOR, Gemma-Emb-0.3B, and Qwen3-Emb-0.6B.
Table~\ref{tab:claim_distribution_clean} reports the verdict category distributions and sizes of the three metric-specific contamination-controlled subsets whose intersection defines this final evaluation set.
Each model is evaluated on this unified contamination-controlled set.

\begin{table}[h]
\centering
\small
\caption{Verdict category distributions of contamination-controlled claims under each similarity metric.}
\label{tab:claim_distribution_clean}
\resizebox{\linewidth}{!}{%
\begin{tabular}{lcccc}
\toprule
\textbf{Categories} & \textbf{METEOR} & \textbf{Gemma-Emb-0.3B} & \textbf{Qwen3-Emb-0.6B} & \textbf{Intersection} \\
\midrule
Supported 
& 50 (12.92\%) 
& 23 (10.55\%) 
& 33 (13.47\%) 
& 19 (12.34\%) \\

Refuted 
& 264 (68.22\%) 
& 157 (72.02\%) 
& 167 (68.16\%) 
& 105 (68.18\%) \\

Not Enough Evidence 
& 73 (18.86\%) 
& 38 (17.43\%) 
& 45 (18.37\%) 
& 30 (19.48\%) \\

\midrule
Total 
& 387 (100.00\%) 
& 218 (100.00\%) 
& 245 (100.00\%) 
& 154 (100.00\%) \\
\bottomrule
\end{tabular}
}
\end{table}

\begin{table}[t]
\centering
\caption{MAFC performance of LLMs on the unified contamination-controlled evaluation set.}
\label{tab:benchmark_average}
\begin{tabular}{lcc}
\toprule
\textbf{LLMs} & \textbf{Accuracy} & \textbf{Macro-F1} \\
\midrule
GPT-5.2
& 51.95\% & \underline{51.66\%} \\
GPT-4o-Mini
& 49.35\% & 39.14\% \\
Gemini-3.0-Pro
& 59.09\% & 50.88\% \\
Gemini-3.0-Flash
& \textbf{63.64\%} & 48.75\% \\
DeepSeek-V3.2
& 55.19\% & \textbf{55.93\%} \\
Qwen3.5-122B-A10B
& \underline{62.34\%} & 47.04\% \\
\bottomrule
\end{tabular}
\end{table}

\paragraph{Results.} The statistics in Table~\ref{tab:claim_distribution_clean} and Table~\ref{tab:benchmark_average} lead to three main observations.
\textbf{Finding 4.1: The evaluation set exhibits noticeable label imbalance.} As shown in Table~\ref{tab:claim_distribution_clean}, all three contamination-controlled subsets are dominated by \textit{Refuted} claims. 
Due to this imbalance, we use Macro-F1 as our primary metric. Future dynamic benchmark construction should more carefully control label distributions. In this regard, mechanisms such as the claim rectification step employed in VERITAS~\cite{veritas} may provide a practical framework for rebalancing claim labels.
\textbf{Finding 4.2: DeepSeek-V3.2 achieves the strongest overall MAFC performance in the contamination-controlled setting.} Specifically, it attains the highest Macro-F1 (55.93\%) on the unified contamination-controlled set.
The smaller GPT-4o-Mini exhibits the lowest MAFC performance, highlighting the impact of model size on fact-checking capability.
\textbf{Finding 4.3: Significant room for improvement remains in MAFC.} All models achieve a Macro-F1 below 56\%, indicating that current approaches remain far from saturated.

\section{Conclusion}

This work re-examines the assumption that dynamic evaluation is inherently contamination-free for multimodal automated fact-checking (MAFC). 
To this end, we investigated four research questions: (\textit{RQ1}) the extent to which existing static and dynamic MAFC benchmarks are contaminated, (\textit{RQ2}) how contamination arises in dynamic benchmarks, (\textit{RQ3}) how such contamination affects MAFC evaluation, and (\textit{RQ4}) how SOTA LLMs perform under contamination-controlled settings.
Our empirical study yields 16 findings, with three key results highlighted below.
First, dynamic benchmarks reduce contamination, but they do not fully eliminate it. 
Second, contamination in dynamic benchmarks arises because even newly published claims are often verifiable using public knowledge available before the cut-off.
Third, contamination significantly inflates evaluation outcomes and can even change model rankings, plausibly because it allows LLMs to shortcut evidence retrieval rather than broadly exploring the evidence space.
These results suggest that contamination can obscure MAFC performance on genuinely unseen claims, prompting us to construct contamination-controlled subsets to reassess SOTA LLMs, further highlighting the substantial room for future improvement.
Beyond evaluation, our contamination detection pipeline can also serve as a data processing tool for decontaminating MAFC benchmarks.

Our work lays the foundation for trustworthy evaluation of MAFC systems in dynamic media environments. We encourage future research to move beyond timestamp-based filtering by adopting more robust controls for claim novelty, advancing MAFC system development, detecting subtle latent contamination beyond observable evidence overlap (for which our similarity-based estimates may be conservative), and expanding empirical studies beyond English to enhance digital media literacy and support integrated information ecosystems for societal well-being.

\newpage
\section*{Acknowledgments}
This work was supported by National Natural Science Foundation of China (No. 62202402),
Guangdong and Hong Kong Universities ``1+1+1'' Joint Research Collaboration Scheme, Project No. 2025A0505000001,
the Early Career Scheme (ECS) from the Research Grants Council of HKSAR (HKBU 22202423),
the General Research Fund (GRF) from the Research Grants Council of HKSAR (HKBU 12203425),
a grant from the Germany/Hong Kong Joint Research Scheme sponsored by the Research Grants Council of HKSAR and the German Academic Exchange Service of Germany (No. G-HKBU208/25),
the Initiation Grant for Faculty Niche Research Areas 2023/24 (No. RC-FNRA-IG/23-24/COMM/01),
Research Cluster Matching Scheme (No. RCMS/24-25/01) of Hong Kong Baptist University,
the grants from the Research Grants Council of HKSAR (HKU 17202325),
the University of Hong Kong (Project 2409100399),
the HKU Faculty Exchange Award 2024 (Faculty of Engineering), 
and Startup Grant (Tier 1) for New Academics AY2020/21 of Hong Kong Baptist University.
\bibliography{main}
\balance
\bibliographystyle{ACM-Reference-Format}

\appendix
\section{Definitions of the Verdict Categories}\label{app:verdict}
Following FEVER~\cite{fever}, 
the ClaimReview2025Q4 benchmark uses three verdict categories for claims:
(i) \textbf{Supported}: The claim is supported by the arguments and evidence presented.  
(ii) \textbf{Refuted}: The claim is contradicted by the arguments and evidence presented.
(iii) \textbf{Not Enough Evidence}: 
The presented evidence is not enough to support or refute the claim.
It applies when the evidence either explicitly indicates that relevant evidence cannot be found or leaves certain aspects of the claim neither supported nor refuted.

\section{Example Prompts}\label{app:prompt}
 
This section presents example prompts.

\tcbset{
  colback=gray!10,
  colframe=black,
  boxrule=0.5pt,
  arc=0pt,
  left=4pt,
  right=4pt,
  top=4pt,
  bottom=4pt,
  breakable
}

\paragraph{Evidence Generation.} We use the following prompt to generate a fact-checking article based on the claim.

\begin{tcolorbox}[breakable]
\textbf{Instructions} Write a fact-checking article to verify the claim.

\textbf{Claim} [CLAIM] 
\end{tcolorbox}

\paragraph{Evidence Extraction.} We use the following prompt to extract evidence from both the oracle article and the LLM-generated article.

\begin{tcolorbox}[breakable]
\textbf{Instructions} You are a precise evidence extraction expert.

\textbf{Claim}  [CLAIM]

\textbf{Task} Extract evidence sentences from the generated text that are directly related to the core factual content of the claim.

\textbf{Rules}\\
1. Only extract content that appears in the generated text.\\
2. Evidence must address the main factual assertion(s) made in the claim.\\
3. Do not infer, summarize, or add information.\\
4. Do not extract sentences that merely restate the claim.\\
5. Avoid duplication.

\textbf{Fact-Checking Article} [FACT-CHECKING ARTICLE]

\textbf{Output} Output strictly in JSON format: \{"Reason": "concise extraction reasoning", "Evidences": [\{"Evidence": "..."\}]\}
\end{tcolorbox}

\section{Impact of LLM Choice on Evidence Extraction}\label{app:robust}

\begin{table}[h]
\centering
\caption{Average contamination scores across different extraction models on ClaimReview2025Q4.}
\label{tab:sim2}
\resizebox{\columnwidth}{!}{
\begin{tabular}{llccc}
\toprule
\textbf{Metric} & \textbf{Extractor} & \textbf{GPT-5.2} & \textbf{Gemini-3.0-Pro} & \textbf{Qwen3.5-122B-A10B} \\
\midrule
\multirow{2}{*}{METEOR} 
& GPT-4o-Mini & 19.15\% & 21.63\% & 23.16\% \\
& GPT-5-Nano  & 18.01\% & 20.95\% & 20.86\% \\
\midrule
\multirow{2}{*}{Gemma-Emb-0.3B}  
& GPT-4o-Mini & 54.12\% & 57.82\% & 60.38\% \\
& GPT-5-Nano  & 53.30\% & 58.18\% & 58.01\% \\
\midrule
\multirow{2}{*}{Qwen3-Emb-0.6B} 
& GPT-4o-Mini & 53.47\% & 56.89\% & 58.48\% \\
& GPT-5-Nano  & 53.57\% & 57.76\% & 56.38\% \\
\bottomrule
\end{tabular}
}
\end{table}

We assess the robustness of our contamination detection pipeline to different LLM extractors. As shown in Table~\ref{tab:sim2}, contamination scores remain consistent when using either GPT-4o-Mini (default) or GPT-5-Nano for evidence extraction.

\begin{figure}[t]
    \centering
    \begin{subfigure}[b]{0.48\linewidth}
        \centering
        \includegraphics[width=\linewidth]{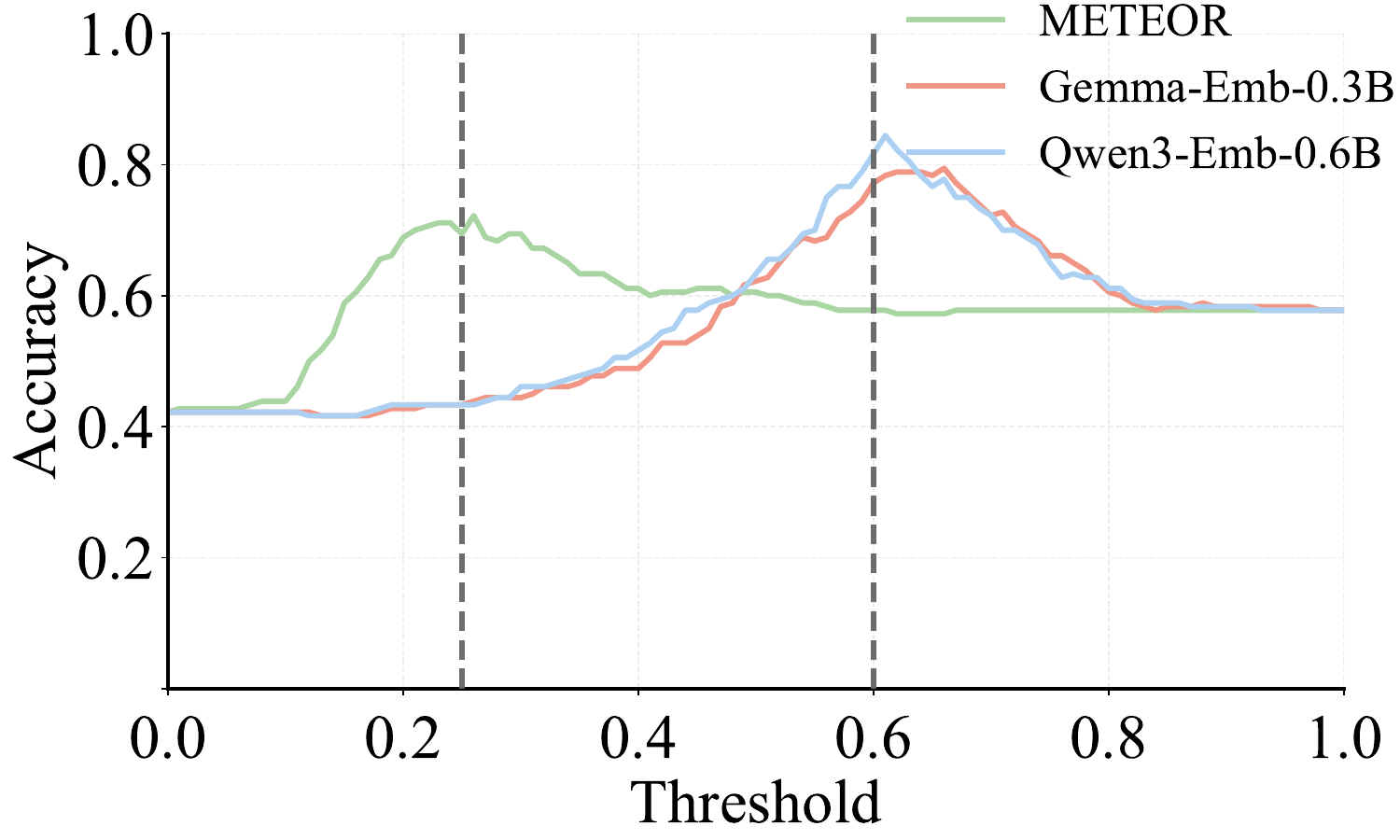}
        \caption{Threshold-sweep analysis.}
        \label{fig:threshold_sweep}
    \end{subfigure}
    \hfill 
    \begin{subfigure}[b]{0.48\linewidth}
        \centering
        \includegraphics[width=\linewidth]{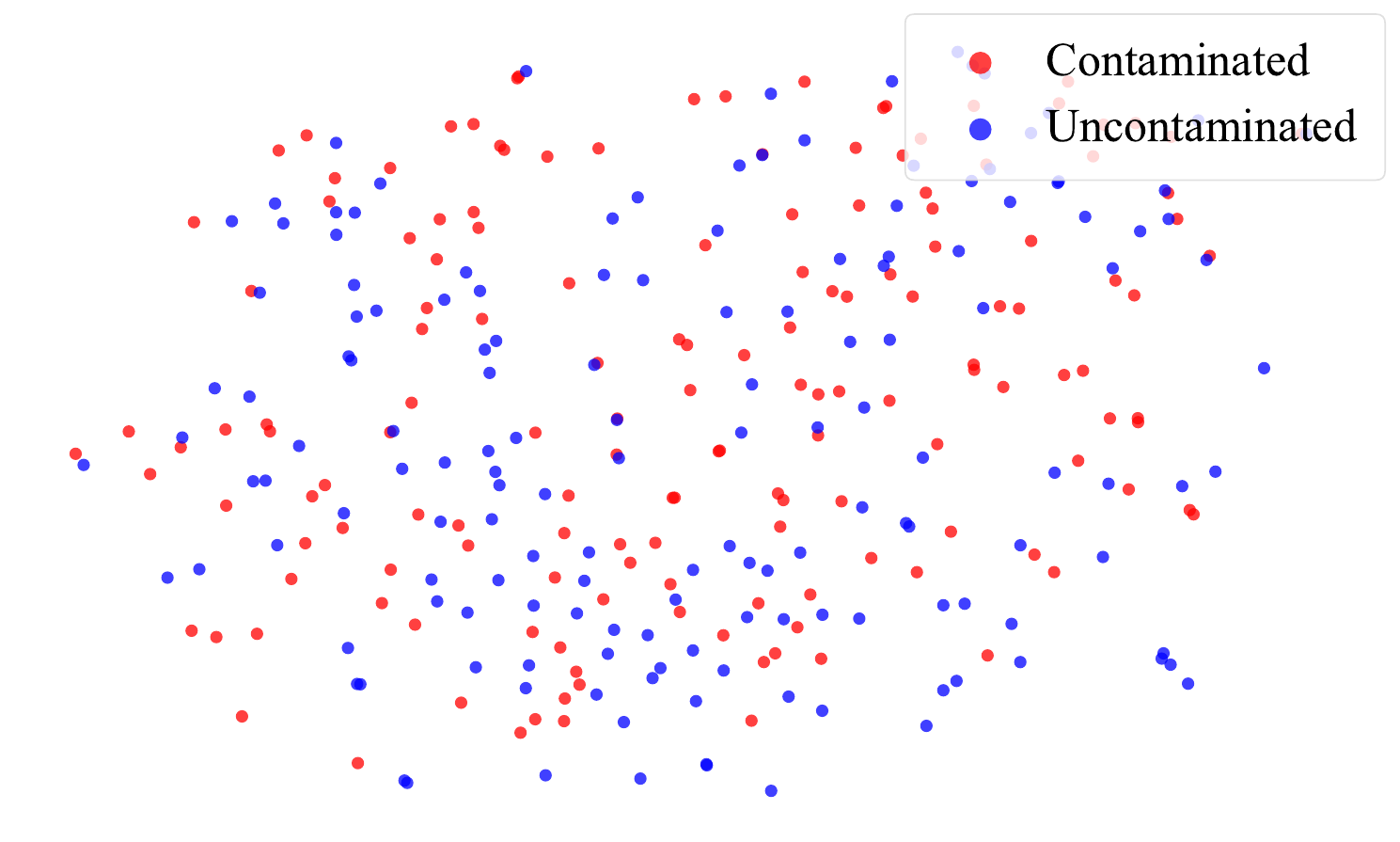}
        \caption{t-SNE visualization.}
        \label{fig:embedding_tsne}
    \end{subfigure}
    \caption{Supporting analyses for threshold selection and subset comparability.}
\end{figure}

\section{Threshold Selection}\label{app:threshold_sweep}
To select the contamination threshold for the semantic similarity metrics, we conducted a threshold-sweep analysis on 180 randomly sampled claims (20\% of ClaimReview2025Q4). Two postgraduate-level annotators with expertise in journalism and/or fact-checking independently labeled each claim as contaminated or uncontaminated. After adjudicating disagreements, we treated the resulting labels as ground truth and evaluated threshold values over $\tau \in [0,1]$.
Fig.~\ref{fig:threshold_sweep} reports the classification accuracy of the contamination labels against these human annotations across threshold values. For the semantic metrics, both curves peak in the range of $\tau = 0.6$--$0.62$, where the accuracy reaches 75.6\% for Gemma-Emb-0.3B and 82.2\% for Qwen3-Emb-0.6B at $\tau = 0.6$. We also performed the same sweep for METEOR, whose peak occurs close to the default AVeriTeC threshold. Therefore, we use $\tau = 0.6$ for the semantic metrics and retain the default threshold of $25\%$ for METEOR in all experiments.

\section{Impact of Non-Contamination Factors}\label{app:disentangle}
A key concern is that performance differences between contaminated and uncontaminated subsets may reflect topical or stylistic variation rather than contamination itself. 
Unlike prior work, both subsets here are drawn from the same ClaimReview2025Q4 sources and time frame, minimizing the likelihood of such confounding factors.
To empirically verify this, we embed each claim using Qwen3-Emb-0.6B (768 dimensions), which captures topical and stylistic features such as language complexity and ambiguity, and compute centroid-based cosine similarities: 88.2\% within the contaminated group, 89.9\% within the uncontaminated group, and 87.9\% across groups.
The high cross-group similarity indicates that the two groups are highly comparable.
Fig.~\ref{fig:embedding_tsne} further visualizes the two groups (downsampled to the same size of 154 claims each) with t-SNE; the points are largely mixed.

\end{document}